\documentclass[10pt,twocolumn,letterpaper]{article}

\usepackage{cvpr}      %

\definecolor{cvprblue}{rgb}{0.21,0.49,0.74}
\usepackage[pagebackref,breaklinks,colorlinks,allcolors=cvprblue]{hyperref}

\def\paperID{*****} %
\def\confName{CVPR}
\def\confYear{2025}

\usepackage[utf8]{inputenc} %
\usepackage[T1]{fontenc}    %
\usepackage{hyperref}       %
\usepackage{url}            %
\usepackage{booktabs}       %
\usepackage{amsfonts}       %
\usepackage{nicefrac}       %
\usepackage{microtype}      %
\usepackage{xcolor}         %
\usepackage{graphicx}
\usepackage{amsmath}
\usepackage{algorithm}
\usepackage{algpseudocode}
\usepackage{listings}
\usepackage{wrapfig}
\usepackage{xcolor}
\usepackage{titletoc}
\usepackage{minitoc}

\lstdefinelanguage{json}{
    basicstyle=\ttfamily\footnotesize,
    numbers=none,
    numberstyle=\tiny\color{gray},
    stepnumber=1,
    numbersep=5pt,
    showstringspaces=false,
    breaklines=true,
    frame=single,
    backgroundcolor=\color{gray!10},
    literate=
     *{0}{{{\color{blue}0}}}{1}
      {1}{{{\color{blue}1}}}{1}
      {2}{{{\color{blue}2}}}{1}
      {3}{{{\color{blue}3}}}{1}
      {4}{{{\color{blue}4}}}{1}
      {5}{{{\color{blue}5}}}{1}
      {6}{{{\color{blue}6}}}{1}
      {7}{{{\color{blue}7}}}{1}
      {8}{{{\color{blue}8}}}{1}
      {9}{{{\color{blue}9}}}{1}
      {:}{{{\color{black}:}}}{1}
      {,}{{{\color{black},}}}{1}
      {"}{{{\color{red}"}}}{1},
}

\title{Audit \& Repair: An Agentic Framework for Consistent Story Visualization in Text-to-Image Diffusion Models}

\author{
Kiymet Akdemir\footnotemark[2] \quad
Tahira Kazimi\footnotemark[2] \quad
Pinar Yanardag \\
Virginia Tech \\
{\tt\small \{kiymet, tahirakazimi, pinary\}@vt.edu}
\\
\texttt{\small \url{http://auditandrepair.github.io}}
}
 
\begin{document}

\twocolumn[{
\maketitle
\begin{center}
    \captionsetup{type=figure}
    \vspace{-2.4em}
\newcommand{\imwidth}{1\textwidth}

\begin{tabular}{@{}c@{}}
 
\parbox{\imwidth} \centering  {\includegraphics[width=0.85\imwidth, ]{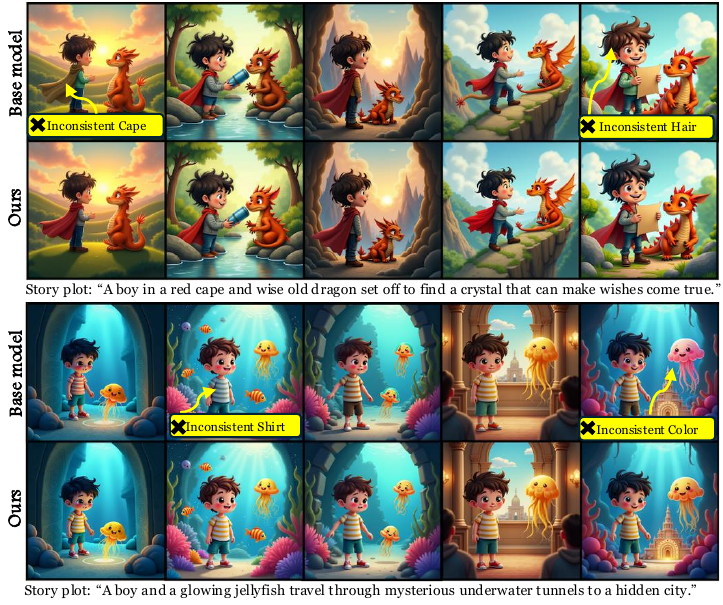}}
\\

\vspace{1em}
\end{tabular}
    \vspace{-3em}
    \captionof{figure}{Given a multi-panel story prompt, our method identifies and corrects visual inconsistencies (e.g., wrong cape color or inconsistent outfit) in character and scene elements across story panels. Compared to a base model (DSD \cite{cai2024diffusionselfdistillationzeroshotcustomized} above), our framework achieves coherent character appearance and narrative consistency throughout the story visualization. } 
    \label{fig:teaser}
\end{center}
}]

\renewcommand{\thefootnote}{\fnsymbol{footnote}}
\footnotetext[2]{Joint first-authors.}
\begin{abstract}
Story visualization has become a popular task where visual scenes are generated to depict a narrative across multiple panels. A central challenge in this setting is maintaining visual consistency, particularly in how characters and objects persist and evolve throughout the story. Despite recent advances in diffusion models, current approaches often fail to preserve key character attributes, leading to incoherent narratives. In this work, we propose a collaborative multi-agent framework that autonomously identifies, corrects, and refines inconsistencies across multi-panel story visualizations. The agents operate in an iterative loop, enabling fine-grained, panel-level updates without re-generating entire sequences. Our framework is model-agnostic and flexibly integrates with a variety of diffusion models, including rectified flow transformers such as Flux and latent diffusion models such as Stable Diffusion. Quantitative and qualitative experiments show that our method outperforms prior approaches in terms  of  multi-panel consistency.  
\end{abstract}

\section{Introduction}

Text-to-image (T2I) diffusion models such as Stable Diffusion \cite{rombach2021highresolution} and Flux \cite{flux2024} have dramatically advanced the fidelity, diversity, and accessibility of image synthesis. These generative models are reshaping creative industries and everyday artistic workflows by enabling the rapid production of illustrations, concept art, and graphics from short text prompts \cite{rombach2021highresolution, saharia2022photorealistictexttoimagediffusionmodels, dalle2}. 

Building on this momentum, story visualization has emerged as a popular task focused on generating a sequence of images that together convey a coherent narrative \cite{li2019storygansequentialconditionalgan, rahman2023make, zhou2024storydiffusion}. Unlike single-image generation, which captures only a single scene, story visualization requires generating a sequence of panels that follow a textual narrative composed of successive sentences or captions. This capability enables compelling applications, including automatic illustration of picture books and comics, rapid generation of storyboards for film production, and interactive visual narratives for video games. However, the main challenge is achieving coherent visual continuity across story panels. Characters, or objects must remain identifiable and stable throughout the narrative. For example, if the character appears in a distinctive striped red t-shirt in the opening frame, the same t-shirt must persist, unaltered, in every subsequent frame in which the character appears, unless noted otherwise by the story prompts. Naively applying a text-to-image model independently to each caption rarely meets this requirement, as subtle drifts in appearance disrupt narrative coherence. Ensuring such continuity is therefore a central, yet unsolved, challenge for generative models \cite{gan, rombach2021highresolution, flux2024}.

Recent approaches such as AutoStudio \cite{cheng2024autostudio}, StoryDiffusion \cite{zhou2024storydiffusion}, and StoryGen \cite{liu2024intelligent} have made progress toward improving consistency in story visualization. However, while they capture general coherence, they often fail to maintain fine-grained consistency in character details and object attributes across panels. Most importantly, these methods treat consistency as an implicit property of the architecture or training loss; once inconsistencies appear, they lack a principled way to detect and repair them at inference time.

In this paper, we present a collaborative multi-agent framework for consistent story visualization that treats the task as an iterative process of refinement across specialized, autonomous components. Rather than relying on end-to-end generation, our framework decomposes the workflow into agents responsible for initializing panels, detecting inconsistencies, refining prompts, and applying visual corrections. A key novelty of our approach is a consistency agent powered by a Vision–Language Model (VLM) that automatically generates textual descriptions of each panel, compares them across the sequence, and flags mismatches in character appearance or object attributes. These flagged inconsistencies are then selectively corrected by other agents, enabling fine-grained updates without the need to re-generate unaffected panels. Our framework is fully model-agnostic and can be integrated with various diffusion backbones—including Flux~\cite{flux2024}, demonstrating substantial improvements in both character and object consistency without any retraining.

\begin{itemize}
\item We propose a collaborative multi-agent framework for story visualization that treats consistency correction as an autonomous, post-generation process across modular agents.

\item We formulate inconsistency detection as a vision–language alignment task, using VLM-generated descriptions to identify fine-grained mismatches in character and object attributes across panels.

\item Our framework introduces the first iterative audit-and-repair loop for multi-frame  consistency that operates independently of the diffusion backbone, and is compatible with both Stable Diffusion and Flux generated stories.

\end{itemize}

\section{Related Work}

\noindent \textbf{Story Visualization}
Recent work has explored diffusion-based methods for story visualization, with growing emphasis on coherence and controllability. StoryDiffusion~\cite{zhou2024storydiffusion} and ConsiStory~\cite{tewel2024consistory} generate multiple frames in parallel using shared self-attention to model narrative consistency. While effective for global coherence, both approaches rely heavily on user-supplied prompts and struggle to maintain consistent character appearances across frames. TaleCrafter~\cite{gong2023talecrafter} and AutoStory~\cite{wang2023autostory} adopt inpainting-based pipelines using LoRA~\cite{hu2021lora} modules, but require finetuning for each LoRA model and depend on structured inputs such as prompts, bounding boxes, or sketches. AR-LDM~\cite{pan2024synthesizing}, StoryGen~\cite{liu2024intelligent}, Make-a-Story~\cite{rahman2023make}, SEED-Story~\cite{yang2024seedstory}, and StoryImager~\cite{tao2024storyimager} follow autoregressive pipelines conditioned on prior frames or captions. While these models promote temporal coherence, they often require dataset-specific training and may compromise visual fidelity or efficiency relative to base diffusion models. OnePrompt-OneStory~\cite{liu2025onepromptonestoryfreelunchconsistenttexttoimage} generates full story sequences from a single paragraph prompt and improves coherence using identity-preserving cross-attention, but offers limited control over per-frame layout and appearance. AutoStudio~\cite{cheng2024autostudio} enhances identity consistency using a dedicated U-Net, though this adds to the model’s complexity. Rather than modifying the generation pipeline or requiring specialized architectures, we take a different perspective: treating consistency as a post-processing problem. Our agent-based framework detects and resolves inconsistencies after generation, offering a modular and model-agnostic solution.

\noindent \textbf{AI Agents for Image Generation.}
Recently, the idea of AI agents has been applied to various image generation tasks, with large language models (LLMs) acting as coordinators. HuggingGPT~\cite{shen2023hugginggptsolvingaitasks} and Visual ChatGPT~\cite{wu2023visualchatgpttalkingdrawing} employ LLMs to select and orchestrate multiple tools, enabling multi-step workflows for image generation and editing. SLD~\cite{wu2023selfcorrectingllmcontrolleddiffusionmodels} introduces an LLM-driven self-correction loop, where outputs are iteratively refined by identifying and addressing prompt mismatches, however their method focuses on improving single-image prompt alignment, rather than multi-frame consistency. Similarly, VideoRepair~\cite{lee2025videorepairimprovingtexttovideogeneration} improves text-video alignment by identifying fine-grained mismatches. For more compositional and structured tasks, GenArtist~\cite{wang2024genartistmultimodalllmagent} and MUSES~\cite{ding2024muses3dcontrollableimagegeneration} coordinate subtasks such as planning, rendering, and editing to generate coherent multi-object or 3D scenes. MM-StoryAgent~\cite{xu2025mmstoryagentimmersivenarratedstorybook} and StoryAgent~\cite{hu2024storyagentcustomizedstorytellingvideo} apply agent-based designs to visual storytelling: MM-StoryAgent integrates text, image, and audio experts to produce storybook videos, while StoryAgent organizes separate agents for writing, image generation, and editing—but requires reference videos of the characters as input, which poses a significant burden for users. Unlike prior work, our system operates across multiple frames to directly address visual consistency. Specialized agents detect inconsistencies, refine prompts, and apply localized edits in an iterative loop without retraining or architectural changes.

\section{Methodology}

\begin{figure*} 
    \centering
    \includegraphics[width=1\linewidth]{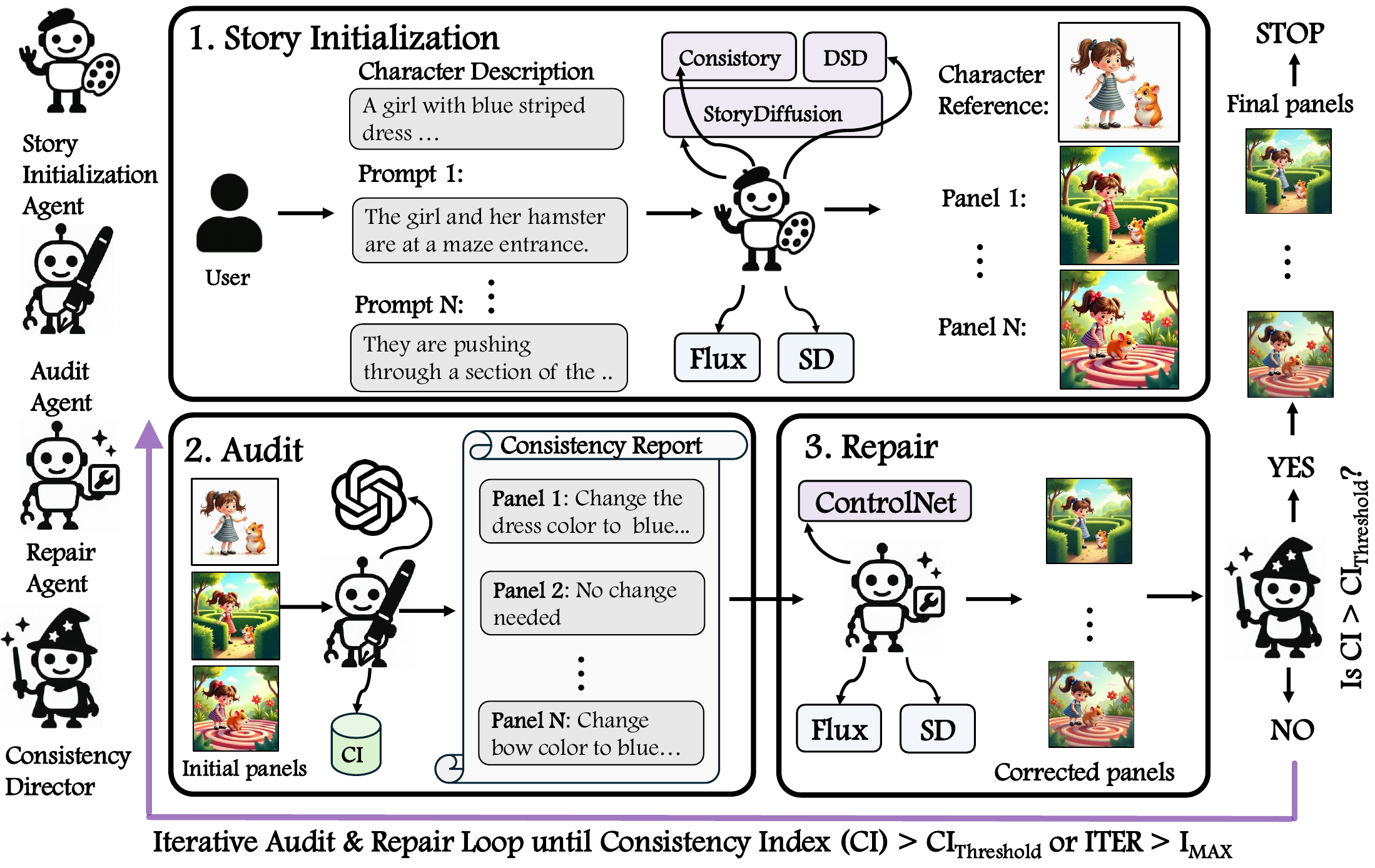}
    \vspace{-5pt}
    \caption{Our framework operates as a collaborative multi-agent system with access to a shared memory that maintains a dynamic Consistency Index (CI), the current panel set, and the latest consistency report. First, Story Initialization Agent, which takes a user-provided sequence of story prompts and character descriptions, and generates initial story panels using a set of off-the-shelf story visualization methods, including both Flux and SD-based models. Once the initial panels are generated, the Audit Agent evaluates each panel using a VLM, updates the CI, and produces a detailed consistency report. This is followed by the Repair Phase, where the Repair Agent applies localized edits to inconsistent panels using editing tools such as Flux-ControlNet. The Consistency Director Agent oversees the entire process, iteratively triggering the Audit and Repair phases until the CI reaches a predefined threshold or a maximum number of refinement iterations is completed.}
    \label{fig:framework}
\end{figure*}
\noindent \textbf{Problem Definition} Given a narrative expressed as $N$ panel—level prompts $\mathcal P={P_1,\dots,P_N}$ and a metadata dictionary $M$ containing details of the characters or objects, the
task is to produce a sequence of images
$\mathcal I={I_1,\dots,I_N}$ such that every recurring character retains all invariant attributes (e.g., hair color, facial features), intentional changes explicitly described in $P_i$ are reflected, and other visual details follow the output of the backbone model.

The naive strategy of feeding each panel description to a T2I backbone in isolation provides no mechanism for propagating character identity across panels, so immutable attributes, such as hairstyle, clothing color, and facial structure, drift unpredictably. Existing story-visualization methods attempt to enforce consistency through training or attention-based mechanisms, but they still treat coherence as a coarse, global property rather than a fine-grained, entity-centric constraint. Without an explicit audit-and-repair loop that separates accidental drift from intentional narrative changes, both naïve generation and attention-augmented models fail to produce sequences that are simultaneously faithful to each prompt and visually coherent over the full story.

\subsection{Multi-Agent Collaborative Design}
To overcome these limitations, we propose a collaborative multi-agent framework that separates generation and correction into modular, specialized agents. Communication is mediated through a shared \emph{memory}, a central data store that continuously tracks the current panel set, the latest consistency report, and the evolving Consistency Index. Each agent autonomously tackles a specific subtask, such as detecting inconsistencies, refining prompts, or applying localized visual edits, within an iterative audit-and-repair loop. This design enables post-hoc refinement of story visualizations, introducing targeted corrections while preserving content that is already correct. Moreover, our framework is model-agnostic and can plug into a variety of T2I diffusion backbones, including Stable Diffusion and Flux, without any training or fine-tuning. Please refer to Fig. \ref{fig:framework} which depicts how our consistency-oriented story-visualization pipeline works. Examples of the prompts used to guide each agent are included in the supplementary material. Our agents are described as follows.

\paragraph{Story Initialization Agent (\(A_{\text{SIA}}\))}
Given the panel-level prompt set \(\mathcal{P} = \{P_1, \dots, P_N\}\) and entity metadata $M$, this agent generates the initial image sequence \(\mathcal{I} = \{I_1, \dots, I_N\}\) using either a story generation model (e.g., StoryDiffusion~\cite{zhou2024storydiffusion}) or an editing-based model (e.g., DSD~\cite{cai2024diffusionselfdistillationzeroshotcustomized}).

For \textit{editing-based methods}, we first use the character descriptions from $M$ to synthesize a single reference image $R$ using a prompt such as \textit{"a girl with a dress and a golden dog"}. Then for each panel, we feed the pair $(R, P_i)$ into the editing model to generate the corresponding image $I_i$.

For \textit{story generation methods}, we prepend the merged character descriptions as an additional prompt to the prompt set, creating an extended prompt set \(\mathcal{P}^* = \{P_0, P_1, \dots, P_N\}\) where $P_0$ contains only character descriptions. The image generated from $P_0$ is used as the reference image $R$, and the rest of the prompt set \(\{P_1, \dots, P_N\}\) is used to generate the full story sequence \(\mathcal{I} = \{I_1, \dots, I_N\}\).

\paragraph{Audit Agent ($A_{\text{Audit}}$)}
The Audit Agent identifies inconsistencies and generates correction instructions. It begins by prompting the VLM to match characters between each panel image $I_i$ and the reference image $R$, using persistent attributes such as clothing, hairstyle, and spatial position to ensure accurate identity mapping. It then prompts the model to describe detailed distinctive character attributes, identify mismatches, and suggest necessary fixes.

To avoid overcorrection, it uses the original prompt $P_i$ to distinguish intentional narrative changes. It further applies a two-step self-verification protocol~\cite{madaan2023selfrefineiterativerefinementselffeedback}: first, checking whether each fix is contextually appropriate, and second, validating that the difference is visually clear and not caused by occlusion or ambiguity.

The agent returns a structured consistency report $\mathcal{C}$ (see supplemental material for an example) and computes the global consistency score:
\[
S_{\text{cons}} = \frac{1}{N} \sum_{i=1}^{N} \text{DINO}(I_i, R), \quad S_{\text{cons}} \in [-1, 1].
\]
where $\text{DINO}(I_i, R)$ denotes the cosine similarity between the DINO embeddings of frame $I_i$ and reference image $R$.
This is linearly rescaled to define the Consistency Index:
\[
\mathrm{CI} = 100\,\frac{S_{\text{cons}} + 1}{2} \;\in [0, 100].
\]

While the Consistency Index is computed globally, edits are applied selectively. The consistency report $\mathcal{C}$ includes frame-level suggestions, and only frames with validated, actionable fixes are passed to the Repair Agent. Before conversion into edits, each suggested fix is also validated for visibility, ensuring that no corrections are proposed for occluded or ambiguous content.
The remaining validated corrections are converted into a list of executable text edits (e.g., \textit{change the hair color of the girl in the dress to black.}) and encoded as refined prompts $\mathcal{P}' = \{P'_1, \dots, P'_N\}$.
\paragraph{Repair Agent ($A_{\text{Repair}}$)}
The Repair Agent receives story frames, along with their corresponding refined prompts $P'_i$ that specify the necessary corrections. It applies localized visual edits using Flux-ControlNet-Union~\cite{flux2024, controlnet}, an editing model that supports image-based conditioning and allows appearance-guided modification of specific visual features.
For each frame, all validated edits are combined into a single semantic prompt to ensure coordinated changes and avoid conflicting updates. After each edit, $A_{\text{Repair}}$ evaluates the outcome and adaptively adjusts the conditioning scale: if the changes are too subtle or not visible, the scale is decreased to encourage stronger edits in future iterations, this information is acquired from the shared memory. If the character’s appearance deviates too much from the reference (indicating over-editing), the scale is increased to reinforce fidelity. Over-editing is flagged by the repair agent when Consistency Index similarity to the reference drops below a threshold, risking unrecoverable character changes in the future iterations. If a valid edit cannot be achieved after several attempts, the panel is skipped and revisited in the next audit. This adaptive strategy enables precise updates while preserving correct content and avoiding full image regeneration.
\begin{figure*}
    \centering
    \includegraphics[width=1\linewidth]{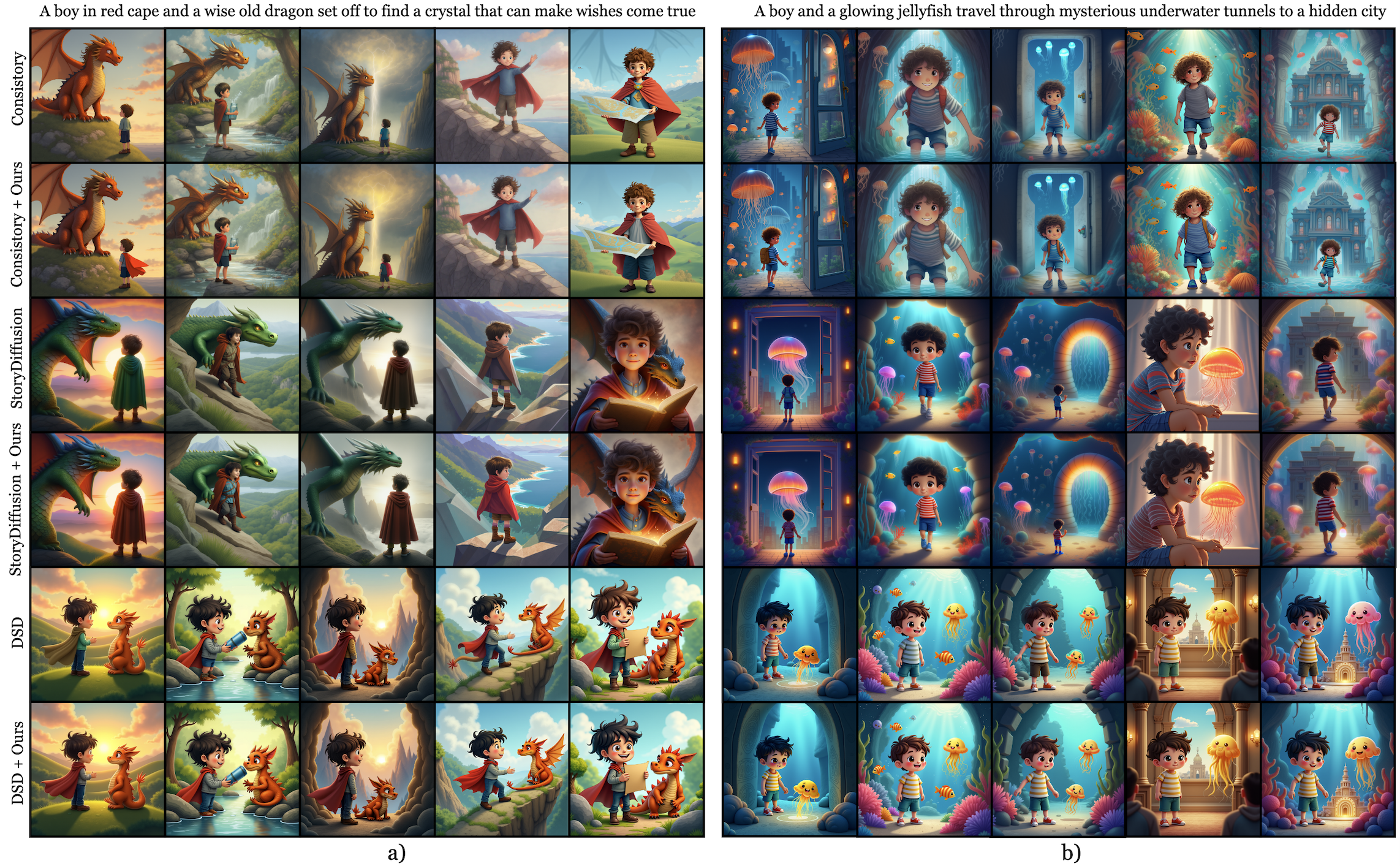}
    \caption{Qualitative results for Audit \& Repair. Our method identifies visual inconsistencies—such as mismatched clothing, character identity  or hairstyle changes—across story panels and refines them using agent-guided editing. The corrected stories preserve narrative coherence and visual consistency, improving character fidelity throughout the sequence.}
    \label{fig:qual-ours}
\end{figure*}
\paragraph{Consistency Director ($A_{\text{CD}}$)}
The Consistency Director oversees the full audit—repair process. After each repair cycle, it requests a new audit from $A_{\text{Audit}}$ and evaluates the updated Consistency Index. If the CI remains below the threshold $\tau$ (default $90$) and the iteration count $T$ is less than or equal to the maximum allowed $T_{\max}$ (default $2$), it initiates another audit—repair loop by calling $A_{\text{Audit}} \rightarrow A_{\text{Repair}}$. Otherwise, it terminates the process and outputs the final, visually consistent story.

Together, these agents operate within a shared-memory, closed-loop control system that casts visual consistency as an \emph{iterative audit–repair cycle}. Guided by the Consistency Director, each detected mismatch is first translated into a lightweight text-level edit and then resolved through disentangled, image-conditioned updates. This strategy (i) \emph{eliminates expensive full-story re-generation}, (ii) \emph{honors any narrative changes explicitly requested by the user}, and (iii)  \emph{remains agnostic} to the underlying diffusion backbone. We set $\tau = 90$ to ensure high visual consistency, and limit the audit-repair loop to $T_{\max} = 2$ iterations to balance correction quality with computational efficiency. These values are selected empirically based on early validation experiments.

\section{Experiments}

To evaluate the effectiveness of our framework, we conducted extensive quantitative and qualitative experiments across multiple metrics and performed detailed ablation studies. Our experiments demonstrate that the proposed framework can be applied to both rectified-flow models (e.g., Flux) and latent-diffusion models (e.g., Stable Diffusion), yielding improvements in visual consistency.

\noindent \textbf{Experimental Setup}  We use GPT-4~\cite{achiam2023gpt} to generate the story prompts, and also as VLM for auditing. Flux-ControlNet-Union~\cite{flux2024, controlnet} is used for localized image repair. The conditioning scale is initially set to 0.37 by default. This scale is later adjusted by the $A_{\text{Repair}}$ to achieve localized edits. The agentic framework is implemented using the AutoGen library \cite{wu2023autogen}. All experiments are conducted on a single NVIDIA A40 GPU. Auditing takes approximately 30 seconds per frame, and editing takes 30 seconds when a repair is triggered. Unless otherwise specified, we report mean and standard deviation over 100 generated stories (in total 700 frames per method). Please see Appendix for an example set of prompts.

\subsection{Qualitative Experiments}
 
\begin{figure*}[t!]
    \centering
    \includegraphics[width=0.9\linewidth]{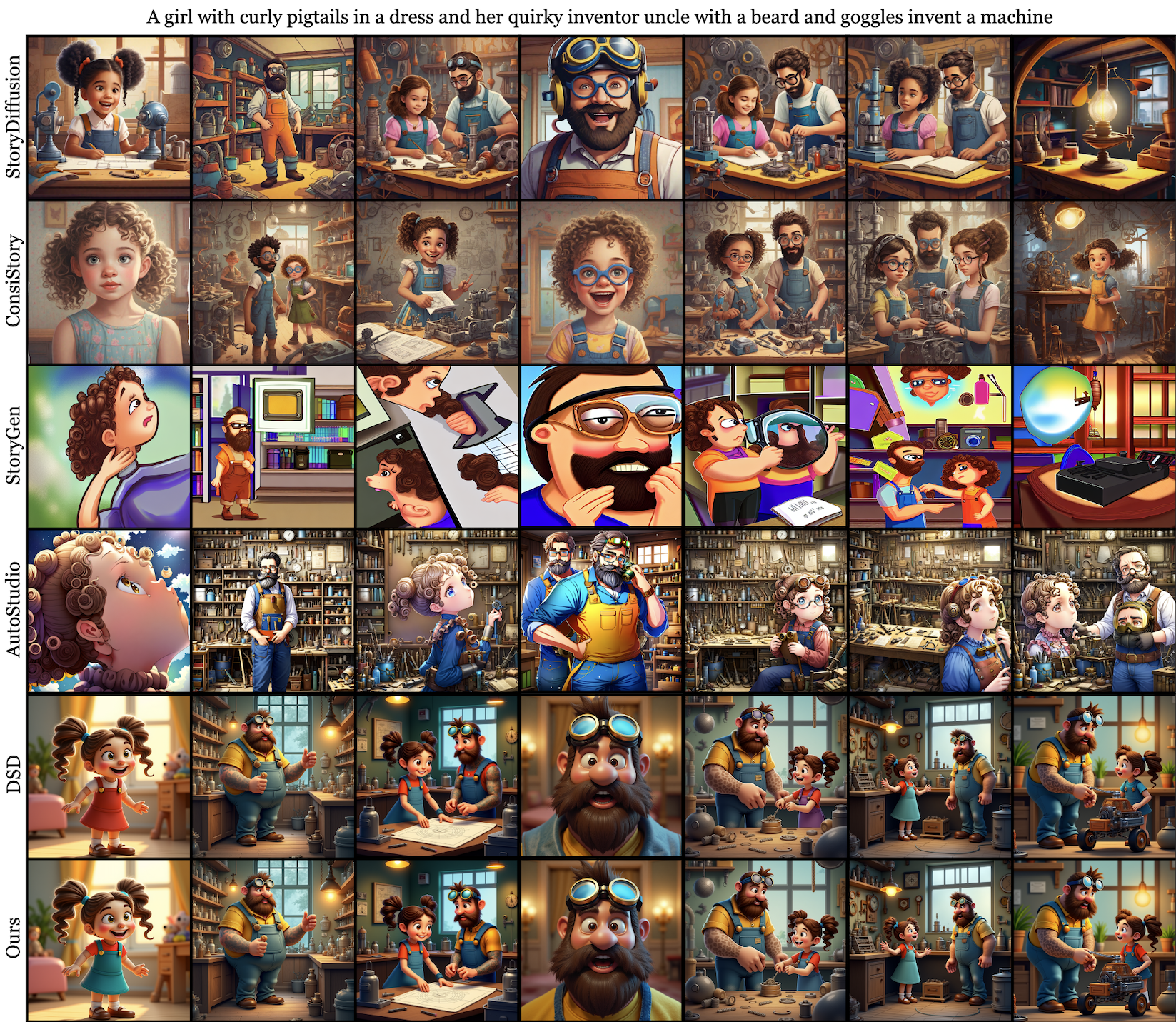}
    \caption{Qualitative comparison of our method with state-of-the-art story visualization methods, including StoryDiffusion, StoryGen, ConsiStory, AutoStudio, and DSD. Our method outperforms existing approaches by preserving consistent visual elements—such as character appearance, clothing, and identity—across all panels. In contrast, prior methods frequently exhibit inconsistencies and blending artifacts that compromise story coherence and the visual identity of characters.}
    \label{fig:qual-comp2}
\end{figure*}

We demonstrate the versatility of our framework across multiple story visualization methods based on both Flux and Stable Diffusion. Specifically, we evaluate three representative models; DSD \cite{cai2024diffusionselfdistillationzeroshotcustomized}, StoryDiffusion \cite{zhou2024storydiffusion}, and ConsiStory ~\cite{tewel2024consistory}. All methods are provided with the same set of story prompts. Fig. \ref{fig:qual-ours} presents comparisons of the original and corrected story sequences produced by our framework. As shown in Fig. \ref{fig:qual-ours}(a), ConsiStory omits key character details such as the red cape, while in Fig. \ref{fig:qual-ours}(b), it removes the backpack and alters the T-shirt color in the final panel. In contrast, our method successfully restores these details. Similarly, StoryDiffusion exhibits inconsistent clothing colors, visible in the first panel of Fig.\ref{fig:qual-ours}(a) and in both the first and fourth panels of Fig.\ref{fig:qual-ours}(b). DSD also suffers from color discrepancies: the cape appears with the wrong hue in Fig.\ref{fig:qual-ours}(a), and the T-shirt color changes between panels in Fig.\ref{fig:qual-ours}(b). Across all cases, our agent-driven correction pipeline identifies and resolves these inconsistencies, yielding visually coherent and consistent story visualizations. 

\begin{figure*}[ht!]
    \centering
    \includegraphics[width=\linewidth]{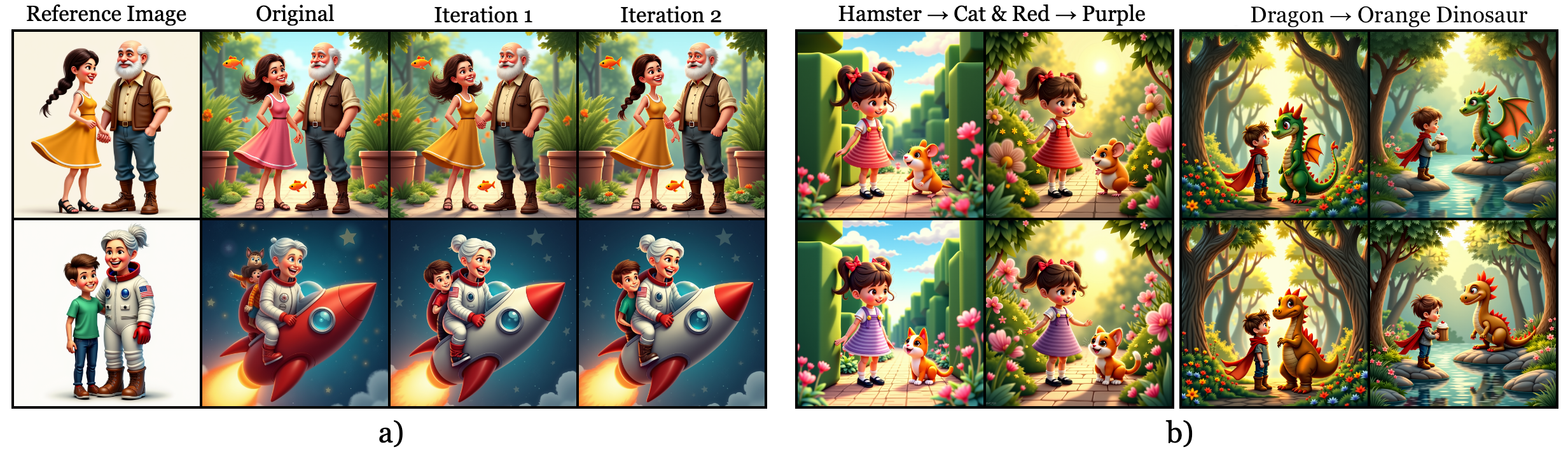}
    \caption{\textbf{a) }Iterative Refinement. Our framework progressively improves consistency across iterations—for instance, correcting dress color in Iteration 1 and hairstyle in Iteration 2, or adjusting a rocket’s design to match the prompt. \textbf{b)} User-in-the-loop Correction. Users can interactively guide edits, enabling both fine-grained adjustments (e.g. dress color) and broader changes (e.g. replacing a hamster with a cat), demonstrating the system’s flexibility and controllability.}
    \label{fig:iteration}

\end{figure*}
\begin{table*}[h]
    \centering
    {\small
    \begin{tabular}{lcccccc}
    \hline
    \textbf{Method} & \textsc{CLIP-I}$\uparrow$ &  \textsc{DINO}$\uparrow$ & \textsc{LPIPS}$\downarrow$ & \textsc{CLIP-T}$\uparrow$ & \textsc{HPS}$\uparrow$ & \textsc{TIFA}$\uparrow$ \\
    \hline
        StoryDiffusion & 0.825$\pm$0.05 & 0.433$\pm$0.10 &  0.494$\pm$0.04 & 0.355$\pm$0.02 & 0.256$\pm$0.03 & 0.661$\pm$0.08  \\
        StoryGen & 0.813$\pm$0.04 & 0.554$\pm$0.10 & 0.631$\pm$0.03 & 0.314$\pm$0.02 & 0.235$\pm$0.03 & 0.494$\pm$0.09\\
        ConsiStory & 0.849$\pm$0.04 & 0.492$\pm$0.10 & 0.513$\pm$0.02 & \textbf{0.359$\pm$0.02} & 0.267$\pm$0.03 & 0.690$\pm$0.09\\
        AutoStudio & 0.782$\pm$0.06 & 0.372$\pm$0.17 & 0.477$\pm$0.18 & 0.338$\pm$0.03 & 0.254$\pm$0.03 & 0.636$\pm$0.09\\
        DSD & 0.845$\pm$0.05 & 0.549$\pm$0.13 & 0.497$\pm$0.05 & 0.351$\pm$0.02 & 0.285$\pm$0.03 & 0.706$\pm$0.08\\
        Ours & \textbf{0.850$\pm$0.05} & \textbf{0.568$\pm$0.15} &  \textbf{ 0.472$\pm$0.07} & 0.351$\pm$0.02 & \textbf{0.319$\pm$0.03} & \textbf{0.713$\pm$0.08}\\
    \hline
    \end{tabular}
    }
    \caption{Average CLIP-I, DINO, LPIPS, CLIP-T, HPS and TIFA scores computed on 100 stories with 7 panels per story, totaling to 700 images per method.}
    \label{tab:quantitative}

\end{table*}
Our framework also accommodates interactive, user-in-the-loop refinements. After the initial panels are generated, a user can provide pinpoint feedback ranging from fine-grained tweaks—such as changing a dress from red to purple—to coarse, semantic replacements, like turning a hamster into a cat or morphing a dragon into a dinosaur (see Fig. \ref{tab:ablation}). Our system parses these natural-language corrections, maps them to localized edit instructions, and re-invokes the diffusion editor only where necessary, leaving unaffected regions intact. Together, these examples underscore the versatility of our framework and its ability to let storytellers steer visualizations toward their exact creative intent.

Next, we compare our method applied (using backbone story visualization model as DSD ~\cite{cai2024diffusionselfdistillationzeroshotcustomized})  with state-of-the-art story visualization baselines, including StoryDiffusion~\cite{zhou2024storydiffusion}, ConsiStory~\cite{tewel2024consistory}, StoryGen~\cite{liu2024intelligent}, AutoStudio~\cite{cheng2024autostudio}, and DSD. Although these models produce plausible image sequences, they often struggle with maintaining character identity and consistency across story panels. As shown in Figure~\ref{fig:qual-comp2}, StoryDiffusion frequently changes key character attributes: the girl's facial features are inconsistent across panels, and the \textit{uncle's clothing shifts between orange and blue overalls}, with noticeable variation in the style and size of \textit{his goggles}. ConsiStory also fails to preserve character identity, altering the \textit{girl's appearance} in nearly every panel and introducing unintended characters (e.g., the sixth frame). StoryGen produces visually unnatural and distorted outputs, with cartoonish exaggerations and incoherent compositions. AutoStudio suffers from blending artifacts (see the fifth panel) and generates incorrect characters, such as the fourth frame which shows \textit{two men} instead of the \textit{described girl and her uncle}. DSD presents inconsistencies in character attire: the uncle wears a sleeveless red shirt in the third frame, while in the fourth, he appears in a jacket rather than overalls. Our method corrects these issues observed in DSD, ensuring that clothing and character identity remain consistent throughout the sequence. In contrast to previous methods, our approach maintains visual consistency across all panels, preserving both the facial features and clothing of the main characters while adapting to pose and scene variations in a way that supports coherent storytelling. See the Appendix for more qualitative examples and visual comparisons.

\subsection{Quantitative Experiments}

We evaluate our method against five state-of-the-art methods—StoryDiffusion~\cite{zhou2024storydiffusion}, StoryGen~\cite{liu2024intelligent}, ConsiStory~\cite{tewel2024consistory}, AutoStudio~\cite{cheng2024autostudio}, and DSD~\cite{cai2024diffusionselfdistillationzeroshotcustomized} and ours (with DSD as the base story visualization backbone). We used the following metrics: CLIP-I~\cite{radford2021learningtransferablevisualmodels}, CLIP-T~\cite{radford2021learningtransferablevisualmodels}, DINO~\cite{caron2021emerging}, LPIPS~\cite{zhang2018perceptual}, TIFA~\cite{hu2023tifa}, and HPS~\cite{wu2023humanpreferencescorev2}. CLIP-I and DINO are image-image similarity metrics, while LPIPS is a perceptual distance metric; all three are used to assess visual consistency across story panels. Specifically, we compute the average pairwise similarity between all frames in a story. CLIP-T, TIFA and HPS measure image-text alignment by evaluating how well each frame corresponds to its associated prompt. As shown in Table~\ref{tab:quantitative}, our method achieves the highest scores on CLIP-I, DINO, indicating improved global and character-level consistency. It also yields the lowest LPIPS score, reflecting improved perceptual similarity across story panels. Compared to DSD, our agentic editing framework yields measurable gains in visual consistency metrics, with an improvement of +\%3.8 in DINO, and a  \%5.0 reduction in LPIPS.  Since computing these metrics on full images often overlooks fine-grained inconsistencies—such as subtle changes in a character’s hair clip across frames—we also present a masked version of the evaluation metrics. Specifically, we use the SAM model ~\cite{kirillov2023segany} to segment and isolate the characters or objects mentioned in the text prompts for each method (see supplementary material for examples of segmented outputs). We then compute the metrics, namely, CLIP-I-FG, DINO-FG, and LPIPS-FG—on these foreground regions to better capture character-level consistency. As shown in Table~\ref{tab:quantitative-fg}, although methods like ConsiStory and StoryDiffusion report competitive scores on the standard CLIP-I metric, their performance drops on CLIP-I-FG. This indicates that their apparent consistency might be attributable to background similarity rather than stable character appearance. In contrast, our method achieves consistently higher performance across all foreground-based metrics, reflecting its ability to preserve semantically meaningful and identity-consistent visual elements across story frames. These results confirm that our agentic framework improves visual coherence in a content-aware manner.

\begin{table*}[t!]
    \centering
    {\small
    \begin{tabular}{lccc}
    \hline
    \textbf{Method} & \textsc{CLIP-I-FG}$\uparrow$ & \textsc{DINO-FG}$\uparrow$ & \textsc{LPIPS-FG}$\downarrow $ \\
    \hline
        StoryDiffusion &  0.814$\pm$0.05 (-1.3\%) &  0.496$\pm$0.10 (+14.5\%) & 0.412$\pm$0.06 (-16.6\%)\\
        StoryGen & 0.810$\pm$0.05 (-0.4\%) & 0.588$\pm$0.13 (+6.1\%)  & 0.470$\pm$0.07 (-25.5\%)\\
        ConsiStory &  0.840$\pm$0.04 (-0.1\%) & 0.579$\pm$0.12 (+17.6\%) & 0.399$\pm$0.06 (-22.2\%)\\
        AutoStudio &  0.811$\pm$0.05 (+3.7\%) & 0.457$\pm$0.20 (+22.8\%) & 0.385$\pm$0.16 (-19.2\%)\\
        DSD &  0.858$\pm$0.05 (+1.5\%) & 0.674$\pm$0.12 (+22.8\%) & 0.337$\pm$0.07 (-32.2\%)\\
        Ours & \textbf{0.860$\pm$0.05} (+1.2\%) & \textbf{0.682$\pm$0.14} (+20.0\%) & \textbf{0.326$\pm$0.08} (-30.9\%)\\
    \hline
    \end{tabular}
    }
    \caption{Average CLIP-I, DINO, and LPIPS scores computed on segmented foreground regions, to further assess character-level similarity without background influence.}
    \label{tab:quantitative-fg}
\end{table*}
Although ConsiStory achieves a higher CLIP-T score than our method, we observe that it often introduces additional characters not mentioned in the prompt (see Figure~\ref{fig:qual-comp2}, sixth panel). This can inflate CLIP-T scores, as the metric measures global semantic alignment without enforcing fine-grained grounding~\cite{hu2023tifa}. Consequently, these hallucinated additions may be rewarded by CLIP-T despite undermining both visual consistency and faithfulness to the intended narrative. To more rigorously assess semantic alignment, we compute the TIFA score~\cite{hu2023tifa} and HPS~\cite{wu2023humanpreferencescorev2} score, a metric designed to test fine-grained faithfulness between text and image. TIFA automatically generates a set of question-answer pairs from the input prompt using GPT-3.5~\cite{brown2020languagemodelsfewshotlearners}, and answers them using UnifiedQA-v2~\cite{khashabi2022unifiedqa}, a visual question answering model. As shown in Table~\ref{tab:quantitative}, our method achieves the highest TIFA and HPS scores across the evaluated benchmarks, demonstrating that it not only improves visual consistency but also enhances faithfulness to the narrative prompts, particularly in compositional accuracy and detail-sensitive elements that CLIP-T may overlook.

\subsection{Ablation Studies}

To evaluate the effectiveness of our framework across different story visualization backbones and show how consistency improves with iterative refinement, we conduct ablation studies as follows.

\noindent \textbf{Ablation on choice of backbone model:} To demonstrate the generality of our editing framework, we apply it to the output of StoryDiffusion~\cite{zhou2024storydiffusion} and ConsiStory~\cite{tewel2024consistory}. As shown in Table~\ref{tab:ablation}, our method consistently improves visual consistency across all metrics. Compared to the original StoryDiffusion output, applying our framework improves DINO by +4.6\%, HPS by +14.5\%, and reduces LPIPS by -1.8\%. For ConsiStory, DINO increases by +4.5\%, HPS improves by +12.7\%, and LPIPS decreases by -2.7\%. These results show that our framework is effective across different generation backbones. Qualitative examples in Figure~\ref{fig:qual-ours} further illustrate corrections in character identity and object attributes.
\begin{table*}[t!]
    \centering
    {\small
    \begin{tabular}{lccc}
    \hline
    \textbf{Method} & \textsc{DINO}$\uparrow$ & \textsc{LPIPS}$\downarrow$ & \textsc{HPS}$\uparrow $ \\
    \hline
        StoryDiffusion &  0.433$\pm$0.09 & 0.494$\pm$0.04  & 0.256$\pm$0.05\\
        StoryDiffusion + Ours & 0.453$\pm$0.09 & 0.485$\pm$0.04 & 0.293$\pm$0.02\\ \hline
        ConsiStory & 0.492$\pm$0.09 & 0.513$\pm$0.02 & 0.267$\pm$0.03\\
        ConsiStory + Ours & 0.514$\pm$0.09 & 0.499$\pm$0.02 & 0.301$\pm$0.02\\\hline
        DSD  &   0.549$\pm$0.13  & 0.497$\pm$0.05 & 0.285$\pm$0.03\\
        DSD + Ours (1st iteration)  & 0.568$\pm$0.15 & 0.472$\pm$ 0.07 & 0.319$\pm$0.03  \\
        DSD + Ours (2nd iteration)  &  \textbf{0.586$\pm$0.13} & \textbf{0.467$\pm$0.05} & \textbf{0.325$\pm$ 0.03} \\

    \hline
    \end{tabular}
    }
    \caption{Ablation results showing consistency improvements after applying our method to StoryDiffusion, ConsiStory and DSD across 100-stories, totaling to 700 panels per method.}
    \label{tab:ablation}

\end{table*}

\noindent \textbf{Ablation on iterative refinement steps:} We evaluate the effect of applying our editing pipeline iteratively. A second audit–repair pass on DSD yields additional improvements: DINO increases by +3.2\%, and LPIPS decreases by -1.1\%. This shows that remaining inconsistencies not resolved in the first pass can be corrected through additional iterations, demonstrating the robustness of our agentic framework. Figure~\ref{fig:iteration} shows how visual consistency improves across iterations. For instance, Iteration 1 corrects the dress color, and Iteration 2 adjusts the hairstyle in Figure~\ref{fig:iteration}(a), top panel. On the Figure~\ref{fig:iteration}(a), bottom panel, Iteration 1 resolves visual artifacts and restores the presence of the boy character with a cape, while Iteration 2 corrects the T-shirt color to match the reference. These examples highlight how our method refines both character consistency and alignment with prompt details.

\begin{table}
  \centering
  \caption{User ratings ($\uparrow$ better) on story alignment and visual consistency, on a 1–5 scale. Our method outperforms others in terms of visual consistency and alignment with narrative.}
  \label{tab:userstudy}
  \begin{tabular}{@{}lcc@{}}
    \toprule
    Method & User-Q1 & User-Q2 \\
    \midrule
    StoryGen & 1.509$\pm$0.88 & 1.465$\pm$0.90 \\
    AutoStudio & 2.210$\pm$1.12 & 2.302$\pm$1.16 \\
    ConsiStory & 2.862$\pm$1.22 & 2.881$\pm$1.29 \\
    StoryDiffusion & 2.279$\pm$1.19 & 2.612$\pm$1.24 \\
    DSD & 3.010$\pm$1.19 & 3.343$\pm$1.08 \\
    Ours & \textbf{3.858$\pm$1.01} & \textbf{3.756$\pm$1.05} \\
    \bottomrule
  \end{tabular}
\end{table}
\subsection{User Study}
To assess human perception of story visualization quality, we conducted a user study with 50 participants recruited via Prolific.com. Each participant was shown a sequence of story panels (144 total per participant) accompanied by the corresponding narrative and asked to rate two aspects on a 1–5 scale: (Q1) how well the visuals maintained consistency across panels, and (Q2) how well the visuals aligned with the narrative. As shown in Table~\ref{tab:userstudy}, our method outperformed all baselines on both criteria. It received the highest ratings for narrative alignment and cross-frame consistency, confirming its ability to generate images that not only follow the story text faithfully but also maintain visual coherence. 
To ensure reliability of the collected ratings, inter-rater agreement metric, measured by Kappa \cite{article}, was calculated over user survey achieving scores (StoryGen: 0.7377, StoryDiffusion: 0.715, Consistory: 0.669, AutoStudio: 0.6704, DSD: 0.679, Ours: 0.707) over six methods. Cohen-Kappa score measures the rate of agreement between the users, with values between 0.6 to 0.8 indicating substantial agreement \cite{article}.

\section{Broader Impact and Limitations}
Our audit-and-repair framework substantially improves consistency across multi-panel stories, lowering the technical barrier for educators, journalists, and independent creators who lack advanced design skills, broadening access to visual storytelling tools. On the other hand, both the underlying VLM and diffusion backbone inherit model biases from their pre-training data—e.g., a tendency to favor mainstream art styles—which can skew similarity estimates and disproportionately flag panels featuring under-represented demographics or unconventional aesthetics.  On a technical limitation, our consistency agent assumes that VLM scores faithfully capture cross-panel coherence. Hallucinations  in these textual descriptions propagate directly to the Consistency Index and may trigger unnecessary—or miss—corrections.  Nevertheless, our framework's audit-and-repair loop consistently improves multi-panel consistency of state-of-the-art methods, providing a strong foundation for further research on consistent story visualization.

\section{Conclusion} In this paper, we presented Audit \& Repair, an agentic framework that approaches story visualization as a coordinated workflow among specialized agents and introduces the first inference-time, VLM-driven mechanism for detecting and repairing cross-panel inconsistencies. By operating entirely at post-generation, our consistency auditor is model-agnostic and can be plugged into any modern T2I backbone, markedly improving narrative coherence without re-training. Comprehensive experiments demonstrate that Audit \& Repair  reduces identity drift while preserving perceptual quality. Future directions include extending the agent hierarchy to temporal domains such as videos. Beyond storyboards and picture books, we believe Audit \& Repair’s modular design provides a principled foundation for scalable, controllable visual storytelling in interactive media and generative design pipelines.
\clearpage

\clearpage
\bibliographystyle{splncs04}
 \bibliography{main}
 
\appendix
\clearpage

\section*{Table of Contents}
\addcontentsline{toc}{section}{Supplementary Material Table of Contents}
\startcontents[appendix]
\printcontents[appendix]{l}{1}{\setcounter{tocdepth}{2}}

\newpage 
\clearpage

\section{Additional Qualitative Results and Comparison}
\begin{figure*}
\centering
    \includegraphics[width=0.8\linewidth]{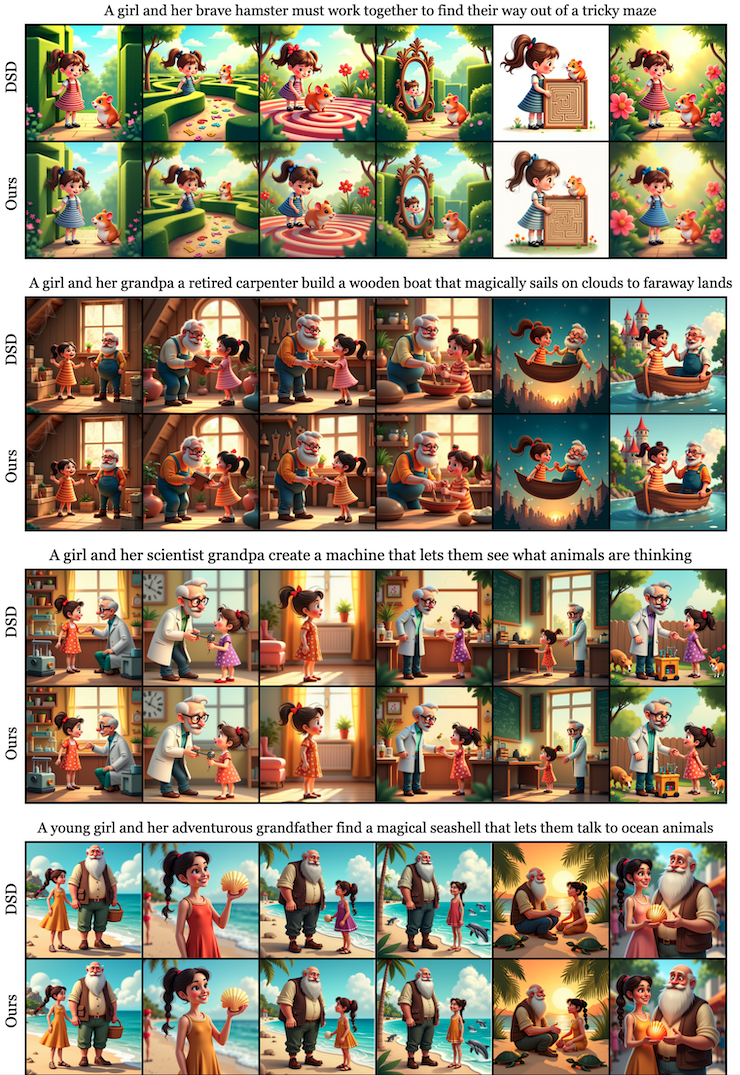}
    \vspace{-0.5em}
    \caption{Inconsistency correction results on the DSD method, illustrated through qualitative examples produced by our agentic framework.}
    \label{fig:qual-appendix}
\end{figure*}

\begin{figure*}
\centering
    \includegraphics[width=0.65\linewidth]{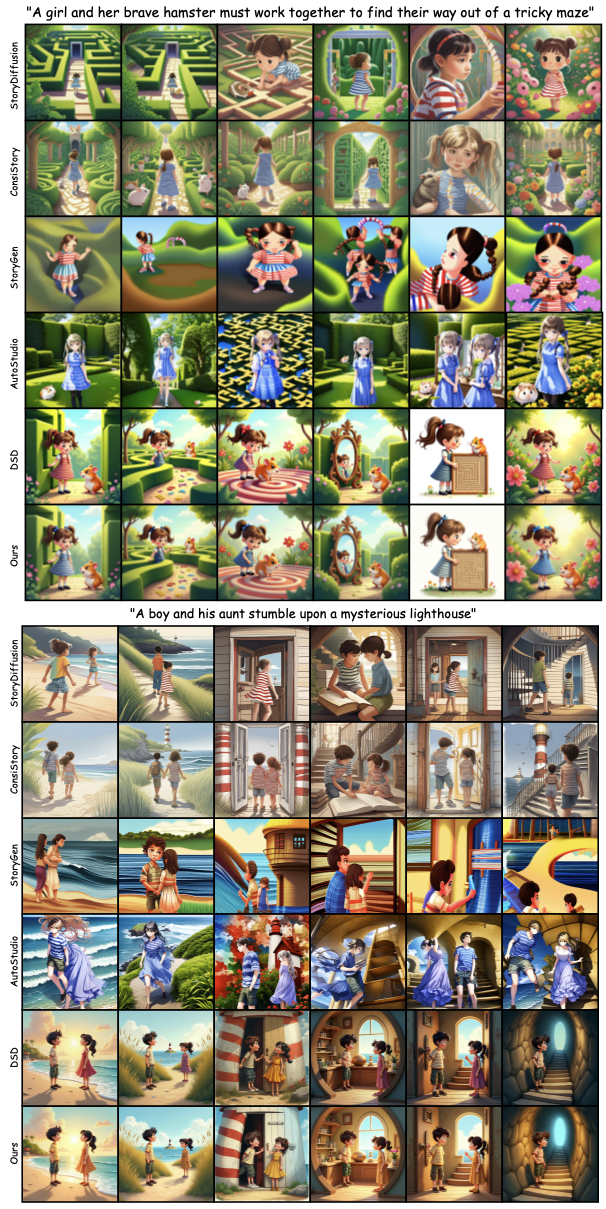}

    \caption{More qualitative comparisons of our method against existing consistency methods.}
    \label{fig:supp-comp}
\end{figure*}

Fig \ref{fig:qual-appendix} and Fig \ref{fig:supp-comp} provide additional qualitative comparisons between our method and several state-of-the-art baselines, including StoryDiffusion \cite{zhou2024storydiffusion}, ConsiStory \cite{tewel2024consistory}, StoryGen \cite{liu2024intelligent}, AutoStudio \cite{cheng2024autostudio}, and DSD \cite{cai2024diffusionselfdistillationzeroshotcustomized}. Figure \ref{fig:qual-appendix} illustrates consistency improvements when using our agentic framework with DSD as the base model, where key elements such as accessories and colors are corrected to remain consistent across panels. Our approach reliably preserves character identity, maintains consistent scene elements (e.g., the maze and hamster), and supports coherent storytelling. In contrast, the baseline methods often exhibit issues such as character drift, inconsistent visual styles, or disjointed narrative flow. Figure \ref{fig:supp-comp} presents further comparisons with other consistency-focused methods. Across all examples, our method produces story sequences that are more semantically accurate and visually coherent, effectively preserving character appearance and relationships throughout the story. These results highlight the strength of our approach in generating consistent and narratively aligned multi-panel visualizations.

\section{Details of User Study}
\begin{figure*}
\centering
    \includegraphics[width=0.75\linewidth]{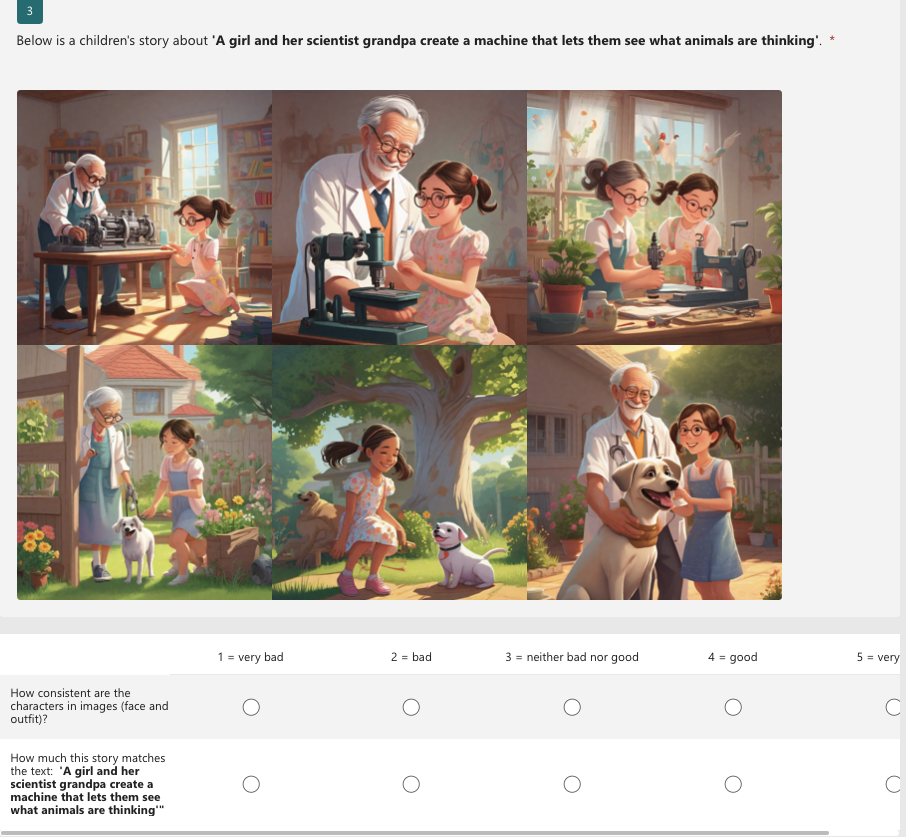}
    \vspace{-0.5em}
\caption{The user study questionnaire includes two evaluation questions: one measuring character consistency and the other assessing story alignment, each rated on a scale from 1 to 5.}
\label{fig:user-study}
\end{figure*}

Figure \ref{fig:user-study} shows an example from our user study survey, where participants were asked to evaluate the visual consistency and story alignment of generated story panels. In the study, we presented frames generated by six different methods, including ours, across five different story prompts. For each story, participants were shown a sequence of six images and asked to rate two aspects on a 1–5 Likert scale: (1) the consistency of characters across panels, and (2) how well the visuals matched the given story text. This structured evaluation allowed us to quantitatively assess both narrative alignment and visual coherence across models in a human-centered manner.

\section{Examples for Segment-based Metrics}

\begin{figure*}
\centering
    \includegraphics[width=0.75\linewidth]{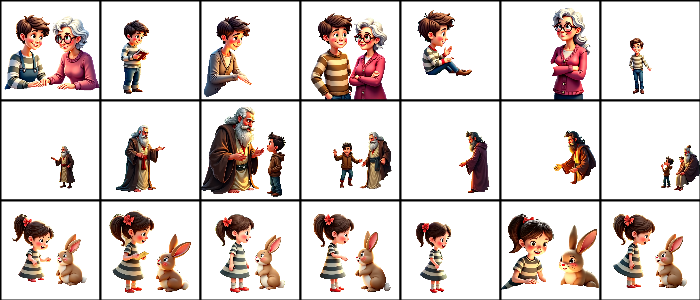}

\caption{Example of segmented foreground images.}
\label{fig:fg}
\end{figure*}

To enable a more reliable comparison, we compute segmented versions of CLIP and other metrics, as standard scores often overemphasize background features and overlook finer details such as changes in clothing or accessories. In Table 2 (main paper), we use the SAM model~\cite{kirillov2023segany} to segment and isolate the characters or objects referenced in the text prompts for each method. Some examples of this process, where segmented characters are extracted and placed against a white background, is illustrated in Figure~\ref{fig:fg}.

\section{Sample Prompts}
Listings \ref{lst:girl_hamster}, \ref{lst:girl_grandpa}, \ref{lst:girl_uncle}, \ref{lst:boy_dragon}, and \ref{lst:boy_jellyfish} showcase sample prompts for various story panels used in the main paper and supplementary material.

\begin{figure*}[t]
\begin{lstlisting}[language=json, caption={Generated story and prompts for "A girl and her brave hamster must work together to find their way out of a tricky maze".}, label={lst:girl_hamster}]
{
  "Main Characters": [
    {
      "Name": "Emily",
      "Description": "A girl with pigtails wearing a striped dress",
      "Category": "girl"
    },
    {
      "Name": "Whiskers",
      "Description": "Small, adventurous hamster",
      "Category": "hamster"
    }
  ],
  "Story": [
    {
      "Image_Prompt": "Emily and Whiskers at a maze entrance.",
      "Location_Description": "Lush green hedges form the complex pathways of the maze, with a clear blue sky overhead and soft sunlight filtering through.",
    },
    {
      "Image_Prompt": "Emily and Whiskers looking at letters on the ground of the maze pathway.",
      "Location_Description": "Lush green hedges form the complex pathways of the maze, with a clear blue sky overhead and soft sunlight filtering through."
    },
    {
      "Image_Prompt": "Emily and Whiskers pushing through a section of the hedge to reveal a secret passage.",
      "Location_Description": "Lush green hedges form the complex pathways of the maze, with a clear blue sky overhead and soft sunlight filtering through."
    },
    {
      "Image_Prompt": "Emily and Whiskers looking at a large ornate mirror in the maze, showing their reflection.",
      "Location_Description": "Lush green hedges form the complex pathways of the maze, with a clear blue sky overhead and soft sunlight filtering through."
    },
    {
      "Image_Prompt": "Close-up of Emily and Whiskers looking thoughtfully at the maze mirror, Whiskers on the maze.",
      "Location_Description": "Maze"
    },
    {
      "Image_Prompt": "Emily and Whiskers standing in the center of the maze, surrounded by flowers and sunlight.",
      "Location_Description": "The center of the maze decorated with blooming flowers and bathed in warm sunlight."
    }
  ]
}
\end{lstlisting}
\end{figure*}

\begin{figure*}
\begin{lstlisting}[language=json, caption={Generated story and prompts for "A girl and her grandpa a retired carpenter build a wooden boat that magically sails on clouds to faraway lands".}, label={lst:girl_grandpa}]
{
  "Main Characters": [
    {
      "Name": "Lily",
      "Description": "a girl with pigtails and a striped dress",
      "Category": "girl"
    },
    {
      "Name": "Grandpa Joe",
      "Description": "an elderly man with a beard and overalls",
      "Category": "man"
    }
  ],
  "Story": [
    {
      "Image_Prompt": "Lily standing and Grandpa Joe showing her a hammer in the attic.",
      "Location_Description": "dusty attic filled with boxes and cobweb"
    },
    {
      "Image_Prompt": "close-up of Grandpa Joe sharing a story with Lily in the attic.",
      "Location_Description": "attic"
    },
    {
      "Image_Prompt": "Lily and Grandpa Joe looking at wooden planks in the workshop.",
      "Location_Description": "a sunny workshop with tools and wood shavings on the floor"
    },
    {
      "Image_Prompt": "Grandpa Joe guiding Lily's hands as she hammers a nail into the boat.",
      "Location_Description": "workshop"
    },
    {
      "Image_Prompt": "Lily in the boat soaring above the clouds, with Grandpa Joe.",
      "Location_Description": "night sky sprinkled with stars above and the silhouette of the town below"
    },
    {
    {
      "Image_Prompt": "Lily steering the boat towards a castle, with Grandpa Joe pointing the way.",
      "Location_Description": "lake"
    }
  ]
}
\end{lstlisting}
\end{figure*}

\begin{figure*}
\begin{lstlisting}[language=json, caption={Generated story and prompts for "A girl with curly pigtails in a dress and her quirky inventor uncle with a beard and goggles invent a machine".}, label={lst:girl_uncle}]
{
  "Main Characters": [
    {
      "Name": "Lucy",
      "Description": "Girl with curly pigtails in a dress",
      "Category": "girl"
    },
    {
      "Name": "Uncle Ned",
      "Description": "Man with a beard and goggles wearing overalls",
      "Category": "man"
    }
  ],
  "Story": [
    {
      "Image_Prompt": "Close-up of Lucy looking up with wide, dreamy eyes.",
      "Location_Description": "room"
    },
    {
      "Image_Prompt": "Uncle Ned walking into the room with greasy hands.",
      "Location_Description": "A cluttered inventor's workshop full of peculiar gadgets and tools."
    },
    {
      "Image_Prompt": "Lucy pointing excitedly at a blueprint on the table.",
      "Location_Description": "A cluttered inventor's workshop full of peculiar gadgets and tools."
    },
    {
      "Image_Prompt": "Close-up of Uncle Ned's face, eyes sparkling with excitement.",
      "Location_Description": "room"
    }
    {
      "Image_Prompt": "Uncle Ned assembling the frame of the machine while Lucy helps.",
      "Location_Description": "A cluttered inventor's workshop full of peculiar gadgets and tools."
    },
    {
      "Image_Prompt": "Lucy and Uncle Ned, with tired eyes, looking at the nearly completed machine.",
      "Location_Description": "A cluttered inventor's workshop full of peculiar gadgets and tools."
    },
    {
      "Image_Prompt": "The completed flying machine gleaming under the light of a single bulb in the workshop.",
      "Location_Description": "A cluttered inventor's workshop full of peculiar gadgets and tools."
    }
  ]
}
\end{lstlisting}
\end{figure*}

\begin{figure*}
\begin{lstlisting}[language=json, caption={Generated story and prompts for "A boy in red cape and a wise old dragon set off to find a crystal that can make wishes come true".}, label={lst:boy_dragon}]
{
  "Main Characters": [
    {
      "Name": "Eli",
      "Description": "A boy with tousled hair and a red cape",
      "Category": "boy"
    },
    {
      "Name": "Zephyr",
      "Description": "A wise old dragon with worn scales",
      "Category": "dragon"
    }
  ],
  "Story": [
    {
      "Image_Prompt": "Eli and Zephyr standing on a hilltop, looking outward.",
      "Location_Description": "Rolling hills under a vibrant sunset sky."
    },
    {
      "Image_Prompt": "Eli holding a parchment map, Zephyr peering over.",
      "Location_Description": "Hilltop with scattered boulders and patches of grass.",
    },
    {
      "Image_Prompt": "Zephyr drinking from the river, Eli refilling a water flask.",
      "Location_Description": "Forest clearing with a clear river bordered by smooth stones."
    },
    {
      "Image_Prompt": "Eli looking up in awe at the towering cliffs, Zephyr's tail in the foreground.",
      "Location_Description": "High, craggy cliffs against a lightening sky."
    },
    {
      "Image_Prompt": "Zephyr assisting Eli as he climbs a steep rocky path.",
      "Location_Description": "Steep rocky pathways with a view of surrounding lands.",
    }
  ]
}
\end{lstlisting}
\end{figure*}

\begin{figure*}
\begin{lstlisting}[language=json, caption={Generated story and prompts for "A boy and a glowing jellyfish travel through mysterious underwater tunnels to a hidden city".}, label={lst:boy_jellyfish}]
{
  "Main Characters": [
    {
      "Name": "Leo",
      "Description": "Boy with wavy hair, in a striped t-shirt and shorts",
      "Category": "boy"
    },
    {
      "Name": "Glimmer",
      "Description": "Luminous and amiable jellyfish",
      "Category": "jellyfish"
    }
  ],
  "Story": [
    {
      "Image_Prompt": "Leo follows Glimmer into a dark underwater tunnel entrance.",
      "Location_Description": "The beginning of a dimly lit and expansive underwater tunnel with rocky walls."
    },
    {
      "Image_Prompt": "Leo and Glimmer swimming with fish around them in a colorful coral-lined tunnel.",
      "Location_Description": "Sunlit coral tunnel bustling with marine life and warm light filtering through."
    },
    {
      "Image_Prompt": "Glowing door with mysterious markings, Leo examining it closely, Glimmer by his side.",
      "Location_Description": "An ancient and mystical section of the underwater tunnel, with engraved walls."
    },
    {
      "Image_Prompt": "Glimmer touching the door, which opens to a glowing city, Leo gazing in awe.",
      "Location_Description": "The magical city with brilliant structures and streets made of luminescent corals and stones."
    },
    {
      "Image_Prompt": "Leo and Glimmer arriving at a grand coral palace with jellyfish guards at the entrance.",
      "Location_Description": "The heart of the city with the grandest building crowned with glowing spires amidst the cityscape."
    }
  ]
}
\end{lstlisting}
\end{figure*}

\end{document}